\documentclass[sigconf,nonacm]{acmart}

\usepackage{graphicx}
\usepackage{booktabs}
\usepackage{amsmath}
\usepackage{amssymb}

\microtypesetup{expansion=true,protrusion=true} 
\setcopyright{none}
\renewcommand\footnotetextcopyrightpermission[1]{}

\begin{document}

\title{When Does Adversarial Refinement Help? A Negative Result and Open Problem in Adapting R3GAN to Time Series Imputation}

\author{Yufeng He}
\affiliation{%
  \institution{The University of Hong Kong}
  \city{Hong Kong}
  \country{China}}
\email{he-yufeng@connect.hku.hk}

\begin{abstract}
Diffusion models and transformers have supplanted GANs for multivariate time series imputation, largely on grounds of GAN training instability.
R3GAN (NeurIPS 2024) removes that instability via regularized relativistic losses with provable convergence, raising a natural question: do stable, modern GANs revive adversarial imputation?
We adapt R3GAN to 1D temporal data with a coarse-to-fine refinement framework and a frequency-domain discriminator, and audit 14 saved configurations across 3 datasets. Because these are heterogeneous single runs, the evidence is descriptive rather than a matched causal ablation.
\textbf{We report a negative result.}
All five saved mean/zero-start configurations improve by 48.4--70.2\%. Among eight eligible non-legacy linear-start configurations, the mean change is $-0.7\%$ (range $-3.0\%$ to $+1.1\%$); a separate $-21.9\%$ legacy logging anomaly is retained for provenance but excluded from that aggregate. In a saved Weather comparison, standalone R3GAN-1D underperforms BRITS by $5.8\times$.
Crucially, we argue the common explanation---that GANs optimize distributional rather than point-wise objectives---\emph{cannot} be the whole story, since diffusion models also optimize distributional objectives yet achieve state-of-the-art imputation.
Our saved reconstruction-weight sweep is consistent with the \emph{adversarial signal} being inert or harmful, but cannot identify its causal contribution; a matched discriminator-removed ablation is the key next experiment.
We frame the precise reason a learned discriminator fails to provide useful refinement gradients---where a learned diffusion denoiser succeeds---as an open problem, and offer practical guidance on when adversarial refinement is worthwhile.
\end{abstract}


\maketitle

\section{Introduction}

Missing values are ubiquitous in multivariate time series (MTS) from domains including healthcare, environmental monitoring, and industrial systems.
Accurate imputation is critical, as missing data degrades the performance of downstream tasks such as forecasting and anomaly detection.

The imputation landscape has evolved rapidly.
RNN-based approaches like BRITS~\citep{cao2018brits} model bidirectional temporal dynamics.
Transformer-based methods such as SAITS~\citep{du2023saits} leverage self-attention for long-range dependencies.
More recently, diffusion models like CSDI~\citep{tashiro2021csdi} and FGTI~\citep{wang2024fgti} achieve state-of-the-art through iterative denoising.
GAN-based imputation, pioneered by GRUI-GAN~\citep{luo2018grui} and GAIN~\citep{yoon2018gain}, showed early promise but has largely vanished from top venues since 2019. The IJCAI 2025 survey~\citep{wang2025survey} notes that GANs ``may face convergence difficulties that can affect imputation accuracy.''

R3GAN~\citep{huang2024r3gan}, a modern GAN with regularized relativistic loss and R1/R2 gradient penalties, achieves provable local convergence for image generation without normalization layers. This makes it a clean testbed for a question the field has not revisited since GANs fell out of favour: \textit{once training instability is removed, can a modern GAN improve time series imputation?}

Our answer is \textbf{largely no}, and we think the negative result is more informative than another marginal positive would have been. We make three contributions:
(1)~\textbf{R3GAN-1D}, a faithful 1D adaptation of R3GAN with a frequency-domain discriminator branch, trained stably in the saved runs;
(2)~a \textbf{reproducibility audit} of 14 saved configurations across 3 datasets that exposes a baseline-strength pattern, retains a documented legacy anomaly, and separates descriptive evidence from a matched causal comparison;
(3)~an \textbf{honest diagnosis and open problem}: the textbook ``distributional-vs-point-wise'' explanation is internally inconsistent (diffusion is also distributional, yet SoTA), while the causal contribution of adversarial discrimination remains to be tested directly.

\section{Related Work}

\textbf{Time series imputation.}
Classical methods include mean/median fill and linear interpolation.
Deep learning approaches span RNN-based (BRITS~\citep{cao2018brits}), attention-based (SAITS~\citep{du2023saits}, ImputeFormer~\citep{nie2024imputeformer}), and generative models.
Among generative approaches, diffusion models (CSDI~\citep{tashiro2021csdi}, FGTI~\citep{wang2024fgti}) now dominate, while GAN-based methods (GRUI-GAN~\citep{luo2018grui}, GAIN~\citep{yoon2018gain}, SSGAN~\citep{miao2021ssgan}) have declined due to training instability.
TSI-Bench~\citep{bansal2024tsibench} provides a comprehensive benchmark showing no single method dominates across all settings.
That diffusion models---also generative, also distribution-matching---now lead imputation is central to our diagnosis (\S\ref{sec:analysis}): it rules out ``generative/distributional objective'' as a blanket explanation for GAN failure.

\textbf{R3GAN.}
\citet{huang2024r3gan} propose a minimalist GAN framework using relativistic paired loss (RpGAN) with zero-centered R1 and R2 gradient penalties. Combined with ResNet-style generators using Fixup initialization (eliminating all normalization layers), R3GAN achieves competitive image generation quality with provable local convergence. We adapt this framework to temporal data and use its stability to test whether training instability is sufficient to explain poor adversarial imputation.

\textbf{Coarse-to-fine imputation.}
Two-stage approaches have been explored in image inpainting, where a coarse network provides initial structure and a refinement network adds detail. We adapt this paradigm to time series, using any simple imputer as the coarse stage and R3GAN as the refinement stage.

\section{Method}

\subsection{R3GAN-1D Architecture}

We adapt R3GAN from 2D image generation to 1D time series, replacing all Conv2d operations with Conv1d and 2D interpolative resampling with 1D lowpass-filtered interpolation using a $[1,2,1]$ kernel.

\textbf{Generator.}
A U-Net-style encoder-decoder with skip connections. Each stage contains inverted-bottleneck residual blocks with grouped convolutions and Fixup initialization (no batch/layer/instance normalization). The generator takes as input the concatenation $[\mathbf{X}_{\text{obs}}, \mathbf{M}, \mathbf{X}_c, \mathbf{z}]$ along the feature dimension, where $\mathbf{X}_{\text{obs}}$ is observed data, $\mathbf{M}$ is the binary observation mask ($M_{tf}=1$ at observed positions and $0$ at missing positions), $\mathbf{X}_c$ is the coarse imputation, and $\mathbf{z} \sim \mathcal{N}(0, I)$ is noise. It outputs a residual $\Delta$:
\begin{equation}
    \hat{\mathbf{X}} = \mathbf{X}_{\text{obs}} \odot \mathbf{M} + (\mathbf{X}_c + \Delta) \odot (1 - \mathbf{M})
\end{equation}
This residual design ensures observed values are preserved exactly and reduces the learning burden---the generator only needs to learn corrections to the coarse estimate.

\textbf{Discriminator.}
Processes complete time series through residual blocks with progressive downsampling. We add a \textbf{frequency-domain branch}: the input's FFT magnitude spectrum is processed by a parallel path, and both features are concatenated before the final logit. This encourages the generator to match spectral characteristics of real data.

\subsection{Training Objective}

Following R3GAN~\citep{huang2024r3gan}, the discriminator uses the regularized relativistic paired loss:
\begin{equation}
    \mathcal{L}_D = \mathbb{E}[\text{sp}(-(D(\mathbf{x}_r) - D(\hat{\mathbf{x}})))] + \frac{\gamma}{2}(R_1 + R_2)
\end{equation}
where $\text{sp}(\cdot) = \log(1 + e^{(\cdot)})$ is the softplus function, and $R_1 = \mathbb{E}[\|\nabla D(\mathbf{x}_r)\|^2]$, $R_2 = \mathbb{E}[\|\nabla D(\hat{\mathbf{x}})\|^2]$ are zero-centered gradient penalties on real and fake samples respectively.

The generator loss combines three terms:
\begin{equation}
    \mathcal{L}_G = \mathcal{L}_{\text{adv}} + \lambda_r \mathcal{L}_{\text{rec}} + \lambda_f \mathcal{L}_{\text{freq}}
\end{equation}
where $\mathcal{M} = \{(t,f) : M_{tf}=0\}$ is the set of \emph{missing} positions (the complement of the observation mask $\mathbf{M}$), $\mathcal{L}_{\text{rec}} = \frac{1}{|\mathcal{M}|}\sum_{(t,f) \in \mathcal{M}} |\hat{x}_{tf} - x_{tf}|$ is the L1 reconstruction loss computed only over those missing positions, and $\mathcal{L}_{\text{freq}} = \||\text{FFT}(\hat{\mathbf{X}})| - |\text{FFT}(\mathbf{X})|\|_1$ encourages spectral fidelity.

We apply cosine learning rate decay, Adam optimizer with $\beta_1 = 0$, $\beta_2 = 0.9$, and maintain an exponential moving average (EMA) of generator weights for evaluation.

\section{Experiments}

\textbf{Datasets.}
We evaluate on three benchmarks spanning different domains and scales: \textbf{Weather} (52K timesteps, 21 meteorological features), \textbf{Electricity} (140K timesteps, 370 client consumption features), and \textbf{AirQuality} (8.7K timesteps, 36 PM2.5 monitoring stations, with 13\% original missing values).

\textbf{Protocol.}
We apply 25\% artificial point-missing (MCAR) and use a 70/15/15 train/val/test split with StandardScaler normalization. Metrics are computed only on artificially masked positions.

\textbf{R3GAN-1D configuration.}
Width=64, 3 stages, 2 blocks/stage, cardinality=16, noise dimension=32. Training: lr=$10^{-4}$, $\gamma$=0.5, $\lambda_r$=20, $\lambda_f$=2, batch size=32, cosine LR decay, EMA. Early stopping based on validation MAE.

\textbf{Baselines.}
BRITS~\citep{cao2018brits} and SAITS~\citep{du2023saits} via the PyPOTS toolkit, plus standard coarse methods (zero fill, mean fill, linear interpolation).

\subsection{Refinement Effectiveness}

\begin{table}[t]
\centering
\caption{Selected saved R3GAN-1D endpoints (MAE$\downarrow$). $\Delta$: relative MAE reduction (positive = improvement). These heterogeneous single runs are descriptive, not a matched repeated-seed comparison.}
\label{tab:refinement}
\small
\begin{tabular}{@{}llccc@{}}
\toprule
Dataset & Coarse Method & Before & After & $\Delta$ \\
\midrule
Weather & Zero fill & 0.728 & 0.228 & \textbf{+68.6\%} \\
        & Mean fill & 0.728 & 0.223 & \textbf{+69.4\%} \\
        & Linear interp & 0.067 & 0.067 & +1.1\% \\
\midrule
Electricity & Zero fill & 0.832 & 0.426 & \textbf{+48.7\%} \\
            & Mean fill & 0.831 & 0.429 & \textbf{+48.4\%} \\
            & Linear interp & 0.164 & 0.165 & $-$0.7\% \\
\midrule
AirQuality & Zero fill & 0.765 & 0.228 & \textbf{+70.2\%} \\
           & Linear interp & 0.151 & 0.152 & $-$0.4\% \\
\bottomrule
\end{tabular}
\end{table}

Table~\ref{tab:refinement} shows representative saved endpoints. Across the full 14-configuration audit, all five zero/mean starts improve by 48.4--70.2\%. The results contain nine linear starts; after excluding one legacy AirQuality run whose log reports zero reconstruction loss in all 200 epochs despite a nonzero configured weight, the eight eligible runs have mean change $-0.7\%$ and range $-3.0\%$ to $+1.1\%$. Thus the saved runs show no consistent gain beyond a plausible coarse fill. The large percentages are improvements over \emph{trivial} baselines and should not be read as competitive performance (cf.\ Table~\ref{tab:sota}).

\begin{figure}[t]
\centering
\includegraphics[width=0.85\columnwidth]{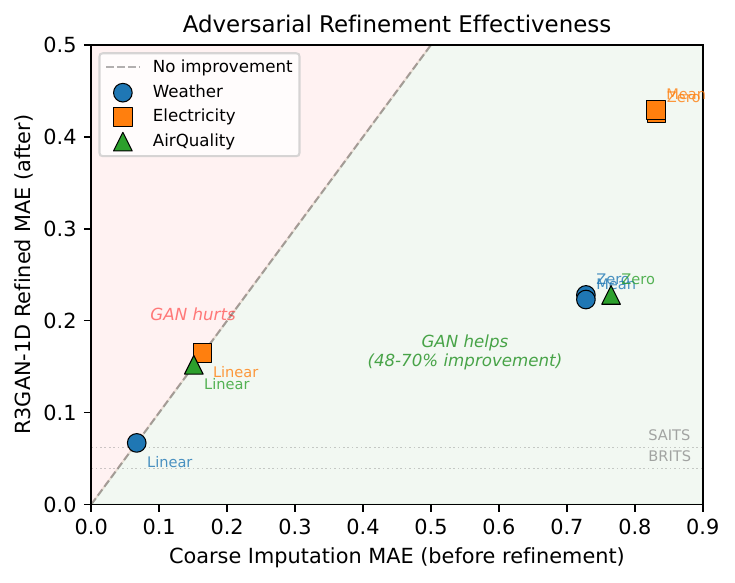}
\caption{Refined MAE vs.\ coarse MAE for the selected endpoints in Table~\ref{tab:refinement}. Points far below the diagonal indicate large improvement. Weak coarse methods (right side) benefit strongly, while linear starts (left side) lie near the diagonal. Dashed horizontal lines mark BRITS/SAITS baselines.}
\Description{Scatter plot comparing coarse-imputation MAE with R3GAN-1D refined MAE. Zero- and mean-fill points lie far below the no-improvement diagonal, while linear-interpolation points lie on or just above it.}
\label{fig:refinement}
\end{figure}

\subsection{Comparison with Established Methods}

\begin{table}[t]
\centering
\caption{Comparison on Weather (rate=0.25, point missing). R3GAN-1D does not match supervised baselines or even linear interpolation.}
\label{tab:sota}
\small
\begin{tabular}{@{}llcc@{}}
\toprule
Method & Type & MAE$\downarrow$ & MSE$\downarrow$ \\
\midrule
BRITS~\citep{cao2018brits} & RNN & \textbf{0.039} & \textbf{0.028} \\
SAITS~\citep{du2023saits} & Transformer & 0.062 & 0.031 \\
Linear interpolation & Simple & 0.067 & 0.079 \\
R3GAN-1D + linear & GAN refine & 0.067 & 0.076 \\
R3GAN-1D standalone & GAN & 0.228 & 0.226 \\
\bottomrule
\end{tabular}
\end{table}

Table~\ref{tab:sota} shows that the saved standalone R3GAN-1D endpoint (MAE=0.228) is 5.8$\times$ worse than BRITS (0.039). The selected linear-start endpoint is numerically unchanged (0.067 $\to$ 0.067), still behind SAITS (0.062). These single endpoints do not establish statistical equivalence; they reinforce that the large saved gains occur only against mean/zero fills that no practitioner would deploy.

\subsection{Ablation Studies}

\textbf{Data augmentation.}
The saved temporal-flip and Gaussian-jitter configurations---standard augmentations in image GANs---degrade imputation quality by $1.8\%$ to $3.0\%$. This is consistent with directional temporal dynamics being harmed by flipping, but repeated matched runs are needed to isolate augmentation effects.

\textbf{Reconstruction weight $\lambda_r$ (key ablation).}
In the saved sweep $\lambda_r \in \{20, 50, 100\}$, higher reconstruction weights do not reduce accuracy below the corresponding coarse endpoint. This pattern is consistent with the adversarial term being inert or harmful for point-wise accuracy, but the configurations are not matched repeated-seed treatments and therefore do not isolate its effect. We return to this boundary in \S\ref{sec:analysis}; a full recon-only comparison remains necessary.

\textbf{Frequency-domain discriminator.}
A saved Weather comparison associates the FFT branch with a marginal $+0.25\%$ change. We treat this as a hypothesis-generating endpoint rather than a component effect.

\section{Analysis: Why, and What Remains Open}
\label{sec:analysis}

\textbf{Why does refinement help weak coarse imputers?}
Mean and zero fill produce outputs that are distributionally implausible---constant or zero stretches where temporal variation should exist.
Transforming these into distributionally realistic signals requires matching global statistics (variance, autocorrelation, spectral density) rather than predicting exact values, which is what an adversarial objective rewards. Hence the large gains in Table~\ref{tab:refinement}---but only relative to baselines that are already far from the data manifold.

\textbf{Why does it fail on strong coarse imputers?}
Linear interpolation already produces locally smooth, temporally coherent signals that sit close to the real-data manifold.
Further improvement requires \emph{point-wise} corrections---the exact deviation from the interpolated value at each missing position.
Empirically (the $\lambda_r$ ablation), the adversarial term supplies no useful gradient for such corrections and can inject ``creative'' variation that raises point-wise error while preserving distributional plausibility.

\textbf{What the usual explanation gets wrong.}
It is tempting to conclude a ``fundamental tension'' between distributional optimization (GANs) and point-wise metrics (MAE/MSE). \emph{We caution against this explanation.}
Diffusion models such as CSDI and FGTI also optimize a distributional/generative objective, yet are state-of-the-art on exactly these point-wise metrics. A blanket ``distributional objectives can't do point-wise imputation'' claim is therefore falsified by existing results. Whatever explains R3GAN-1D's failure must be specific to \emph{adversarial} training, not to generative modelling in general.

\textbf{An open problem.}
We can only sharpen, not settle, the question. A working hypothesis: a diffusion model is trained by a per-position denoising regression conditioned on observed values, so its learning signal is itself point-wise and observation-aware; a GAN discriminator instead scores \emph{global} plausibility and, once the input is already on-manifold, backpropagates little position-specific information. If correct, the deficit is in the \emph{form of the learning signal} (global discrimination vs.\ conditional per-position regression), not in distribution-matching per se. Testing this---e.g., by replacing the discriminator with a conditional, per-position critic, or by interpolating between adversarial and denoising-score objectives---is, to our knowledge, open. We offer R3GAN-1D and this framing as a concrete starting point.

\textbf{Practical guidance.}
From the descriptive saved-run evidence we recommend:
(1) for accuracy-critical imputation, prefer supervised or diffusion methods (BRITS, SAITS, CSDI);
(2) treat GANs as \emph{coarse reconstructors} only when starting from trivial baselines, or as data augmenters generating plausible synthetic samples;
(3) hybrid designs---e.g., GAN-generated augmentation to improve a supervised or diffusion imputer---may combine distributional diversity with point-wise precision, but this remains to be demonstrated.

\section{Limitations}

Our study is deliberately scoped and has clear limitations.
First, the reproducibility audit covers 14 heterogeneous saved configurations, each with one stochastic run and without a complete recorded seed provenance. The selected table entries are not repeated trials or a controlled coarse-method ablation. One legacy $-21.9\%$ linear-start run is retained but excluded from the aggregate because its log records zero reconstruction loss in all 200 epochs despite a nonzero configured weight. Consequently, the reported ranges and means are descriptive, with no uncertainty interval or causal interpretation.
Second, and most important, \textbf{we do not include a matched recon-only (discriminator-removed) ablation}; the $\lambda_r$ sweep cannot establish whether the discriminator helps, hurts, or is inert. A multi-seed comparison sharing masks, splits, scaling, initialization, and generator stochastic streams is the single most valuable next experiment and could confirm or falsify the diagnosis.
Third, our supervised comparison (Table~\ref{tab:sota}) is reported on Weather, and the diffusion methods central to our argument (CSDI, FGTI) are cited rather than re-run in our protocol; an in-protocol diffusion baseline across all three datasets is the natural complement to the recon-only ablation, and we leave both to follow-up work.
Fourth, we evaluate reconstruction MAE/MSE only, not downstream-task utility; a distributionally richer imputer could in principle help downstream tasks even at equal MAE.
We flag these so the negative result is read for exactly what it supports: in the eligible saved linear-start runs, the full system shows no consistent point-wise gain over a competent coarse imputer; the discriminator's causal contribution remains unresolved.

\section{Conclusion}

We adapted R3GAN to time series imputation and report a scoped, descriptive negative result: saved endpoints show large gains over distributionally trivial coarse fills but no consistent gain among eight eligible linear-start configurations, while a selected Weather comparison remains behind supervised baselines.
Diffusion's success shows that ``distributional-vs-point-wise'' cannot be a complete explanation. The discriminator's causal contribution is still unresolved, and the matched recon-only experiment specified in our public protocol is designed to answer it.
We hope that publishing the complete saved-run boundary, including the legacy anomaly, prevents overinterpretation and turns ``GANs just don't work here'' into a falsifiable experimental question.

\textbf{Code:} \url{https://github.com/he-yufeng/adversarial-refinement-imputation}.

\bibliographystyle{ACM-Reference-Format}
\bibliography{references}

\end{document}